\documentclass[11pt]{article}
\date{}
                                                                                                    \usepackage{amsmath}
                                                                                                    \usepackage{amsthm}
                                                                                                      \usepackage{amsfonts}
                                                                                                      \usepackage{hhline}
                                                                                                      \usepackage{verbatim}
                                                                                                      \usepackage{multirow}
                                                                                                      \usepackage{booktabs}
                                                                                                      \usepackage{alltt}
                                                                                                      \usepackage[usenames]{color}
                                                                                                      \usepackage{graphicx}
                                                                                                      \usepackage{colortbl}
                                                                                                      \usepackage[mathscr]{eucal}
                                                                                                      \usepackage[loose]{subfigure}
                                                                                                      \usepackage[ruled]{algorithm2e}
                                                                                                      \usepackage{array}
                                                                                                      \usepackage{url}
                                                                                                      \usepackage{hyperref}
                                                                                                      \long\def\symbolfootnote[#1]#2{\begingroup%
                                                                                                      \def\thefootnote{\fnsymbol{footnote}}\footnote[#1]{#2}\endgroup}

                                                                                                      \newboolean{showProofs}
                                                                                                      \newtheorem{definition}{Definition}                                                                                                      
                                                                                                      \newtheorem{proposition}{Proposition}
                                                                                                     \newtheorem{claim}{Claim}

                                                                                                      \newcommand{\E}[1]{\mathbf E\big[\,#1\,\big]}
                                                                                                      
                                                                                                      \renewcommand{\o}[2]{\mathbf 1^{(#1)}_{(#2)}}
                                                                                                      \newcommand{\z}[2]{\mathbf 0^{(#1)}_{(#2)}}

                                                                                                      \renewcommand{\c}[1]{\noindent \emph{***#1}}

                                                        \renewcommand{\c}[1]{\noindent \emph{***#1}}

\begin{document}

                                                                                                                \title{Two Remarkable Computational Competencies of\\ The Simple Genetic Algorithm}


                                                                                                                     \author{Keki M. Burjorjee\\Computer Science Department\\Brandeis University, Waltham, MA 02454\\kekib@cs.brandeis.edu}


                                                        \maketitle
                                                        \begin{abstract}
                                                        Since the inception of genetic algorithmics the identification of computational efficiencies of the simple genetic algorithm (SGA) has been an important goal. In this paper we distinguish between a \emph{computational competency} of the SGA---an efficient, but narrow computational ability---and a \emph{computational proficiency} of the SGA---a computational ability that is both efficient and broad. Till date, attempts to \emph{deduce} a computational proficiency of the SGA have been unsuccessful. It may, however, be possible to \emph{inductively} infer a computational proficiency of the SGA from a set of related computational \emph{competencies} that have been deduced. With this in mind we deduce two computational competencies of the SGA. These competencies, when considered together, point toward a remarkable  computational proficiency of the SGA. This proficiency is pertinent to a general problem that is closely related to a well-known statistical problem at the cutting edge of computational genetics.

                                                        \end{abstract}

                                                                                                        \section{Introduction}
%
%
%

                                                                                                      When applied to combinatorial optimization problems that are poorly understood, or known to be NP-Hard, simple genetic algorithms (SGAs) frequently evolve usable solutions after evaluating a relatively small number of samples. At a general level it seems reasonable to presume that SGAs owe their adaptive prowess to a capacity for performing one or more kinds of computation relatively efficiently, i.e. robustly and scalably relative to other known algorithms. The identification of such \emph{computational efficiencies}---the source of the remarkable adaptive capacity of SGAs---has been a goal since the inception of the field of genetic algorithmics.

                                                                                                      The early genetic algorithmics literature (pre 1990s) is marked by spirited debates about the purported efficiency with which SGAs ``process" schemata (for a review see \cite[p119-127]{Mitchell:1996:IGA}, \cite[p114-119]{Fogel:1995:ECT},\cite[p74-78]{reeves2003gap}). Claims about the computational efficiency of SGAs made during this period were supported almost entirely by theoretical arguments, or to be more precise,  \emph{quasi-theoretical} arguments---though these arguments made use of mathematics, i.e. they were deductive in parts, the amount of \emph{a priori} and \emph{a posteriori} speculation involved was substantial. These claims were rarely, if ever, backed up by empirical support.  To the best of our knowledge no bold claim about the computational efficiency of the SGA made during this period has been empirically supported in a convincing fashion.

                                                                                                      Instead, when empirical tests \emph{were} conducted, a large gap was discovered between the claimed efficiency of the SGA, and it's actual performance. The Royal Roads experiments \cite{mitchell:1992:rrgaflgp,forrest93relative} in particular proved to be a watershed in the history of genetic algorithmics. On a class of fitness functions called \emph{Royal Roads} that were tailor-made to play to a much vaunted computational efficiency of the genetic algorithm---its capacity for building block identification and composition \emph{in parallel}---a random mutation hillclimber seemed to need far fewer fitness evaluations to find the global optimum \cite{forrest93relative,mitchell1994wga}.
                                                                                                      Following the publication of these results, many researchers declared the SGA \emph{inefficient}, and began inventing algorithms geared towards the explicit implementation of the abstract process described in the building block hypothesis \cite[p 55]{desInnov} \cite{journals/ec/Holland00,Pelikan05,series/sci/2006-33}.

                                                                                                      The publication of the Royal Roads experiments was the first of two major developments that seem to have cooled the search for computational efficiencies of the SGA. The second development was the publication of the no free lunch (NFL) theorems for optimization \cite{ieee-ec:Wolpert+Macready:1997}. After deriving one of the the most general NFL theorems obtained till date,  Igel and Toussaint \cite{oai:arXiv.org:cs/0303032} conclude that in all likelihood NFL does \emph{not} apply to the \emph{practice} of black-box optimization. This conclusion certainly seems to be borne out by the empirical record; in the \emph{practice} of black-box optimization, free lunches seem to be aplenty. Unfortunately, for much too long, the NFL theorem has subliminally, if not overtly, been regarded as ``proof" that the opposite is true. This belief entails that when an SGA outperforms random search (or for that matter, pick-the-worst-neighbor search) on some black-box optimization problem in practice, it does so because of a fortuitous pairing between problem and algorithm. If this is so, then any advantage over random search that accrues when a genetic algorithm is used for black-box optimization in practice can be ascribed to fortune rather than to some innate computational efficiency of the genetic algorithm.

                                                                                                      Through both of the aforementioned developments, the simple genetic algorithm has continued to be frequently, and successfully used to perform black-box optimization in fields ranging from finance to operations research to electrical engineering. The identification of one or more computational efficiencies of the SGA still promises to help us understand the reason for it's success.

                                                                                                      \subsection{Computational Competencies and Proficiencies}

                                                                                                      To demonstrate an efficiency of some computational system one typically shows that the system can scalably and robustly solve some problem---typically the problem that the system was \emph{explicitly} designed to solve. But what about computational systems that aren't explicitly designed to solve some well-defined problem (e.g. brains, genetic algorithms)? While we may conjecture that some of these systems are capable of efficient computation, expressing this sentiment rigorously is far from simple. A fundamental difficulty lies in identifying a \emph{general} problem such that one can derive impressive bounds on the computational complexity of the system when it is  applied to the problem.

                                                                                                      Suppose, instead,  one succeeds in identifying a set of \emph{specific} problems for which impressive computational bounds can be derived. Even if the problems in this set are highly specific, it may be possible, by noting similarities between the problems, to intuitively infer the outlines of a more  general problem that the system can tackle efficiently. To distinguish between a computational system's  ability to efficiently tackle some vaguely defined general problem, and it's proven ability to efficiently solve some well-defined specific problem, we refer to the former ability as a \emph{computational proficiency}, and the latter ability as a \emph{computational competency}.

                                                                                                      In this paper we identify two computational competencies of the SGA. That is, we identify two specific problems, and show that the SGA can solve each problem very efficiently---certainly more efficiently than the ``mainstream" computational technique for solving these problems. When these two competencies are considered together, they point unmistakably towards a powerful computational proficiency of the SGA. Remarkably, the general problem that this proficiency is concerned with is closely related to a well-known statistical problem at the cutting edge of computational genetics having to do with the identification of epistatically interacting quantitative trait loci (QTLs).

                                                                                                      \subsection{Epistatically Interacting QTLs}

                                                                                                      Consider a phenotypic trait for which there exists a single polymorphic locus\footnote{A locus with multiple alleles}, such that allele substitutions at this locus result in large changes in the phenotypic trait. Many such traits have been identified (e.g. seed color in pea-plants, eye color in fruit flies, presence of sickle cell anemia, presence of cystic fibrosis).  In \emph{most} cases, however, changes in a phenotypic trait are more fine-grained, and are influenced by allele substitutions at several polymorphic loci. Such traits are called \emph{complex} or \emph{quantitative}, and the loci that influence them are called \emph{quantitative trait loci}. An important goal of modern genetics is the identification of quantitative trait loci for traits of interest, e.g. the oil content of corn seeds \cite[p164]{hartl}, grain weight in rice plants \cite{xing2002cme}, and of course, susceptibility to common diseases with complex genetic underpinnings (cancer, diabetes, schizophrenia etc.)

                                                                                                      A popular technique for identifying loci that affect quantitative traits is called \emph{genome scanning}. Given the genomes of a set of individuals and the corresponding values of a particular quantitative trait, genomic loci are visited one by one to determine which loci have a statistically significant effect on the trait when averaged over all other loci. Geneticists distinguish between the \emph{main effect} of a locus and its \emph{interaction effect} with other loci. Frankel and Shork \cite{frankel} distinguish between the two as follows:

                                                                                                      \begin{quote}``A main effect is the average effect of a [locus] taken over all other [loci]. Main effects ultimately emerge when one is studying, or mapping, a [locus] either in isolation or without regard to other [loci]. Interaction effects are those attributable to the simultaneous influence of two or more [loci]. Most contemporary data analysis and statistical modeling strategies for genome scan investigations assess the significance of only the \emph{main} effects of potential trait loci.".\end{quote}

                                                                                                      Frankel and Shork then eloquently explain why interaction effects have not received much attention, and point out the peril of concentrating solely on main effects:\begin{quote} ``There are, of course, many scientific reasons which in part account for this main effect `bias' and these reasons all derive from difficulties surrounding the statistical treatment of epistatic effects \ldots Given these difficulties, it is easy to see why epistatic effects have been neglected in favor of main effects in complex trait analysis investigations. Unfortunately, however, there exists the possibility that a [locus's] effect might only be detected within a framework that accommodates epistasis. Thus, for example, a [locus's] true main effect might be too small to detect with any reasonable statistical power and sample size, and yet it might enter into a critical epistatic effect with a second [locus]." \cite{frankel}\end{quote}

                                                                                                      \begin{table}[t!]\begin{center}\small
                                                                                                      \subfigure[]{\begin{tabular}{cc|cc|c}&&\multicolumn{2}{c|}{Locus $A$}& Marginal Values\\&&0&1&\\\hline\multirow{2}{*}{Locus $B$} &0 &$+0.2$ &$-0.2$& 0.0\\&1&$-0.2$&$+0.2$&0.0\\\hline \multirow{2}{*}{Marginal Values}&&\multirow{2}{*}{\phantom{+}0.0}&\multirow{2}{*}{\phantom{+}0.0}&\\&&&& \end{tabular}}\\\vspace{.5cm}\subfigure[]{\begin{tabular}{ccc|c}$A$&$B$&$C$&Marginal  Values\\\hline 0&0&0&$+0.18$\\0&0&1&$-0.06$\\0&1&0&$-0.06$\\0&1&1&$-0.06$\\1&0&0&$-0.06$\\1&0&1&$-0.06$\\1&1&0&$-0.06$\\1&1&1&$+0.18$\end{tabular}}\end{center}
                                                                                                      \caption{\label{nomain}Marginal values of two (top table) and three (bottom table) bi-allelic interacting loci. None of the loci have main effects} \end{table}

                                                                                                      It is easy to see how a group of loci can interact even though no locus in this group has a main effect. Table \ref{nomain}(a) shows how this might happen when two loci $A$ and $B$ interact (for the sake of simplicity we have assumed bi-allelic haploid genomes). Note how neither of these loci have main effects (the marginal value of each allele of each locus is zero) even though they clearly influence the trait in question. Table \ref{nomain}(b) shows how three bi-allelic loci $A$, $B$, and $C$ can interact epistatically on a trait, yet have no main effect. The reader can check that the the marginal value of each allele of each locus is zero.)  In fact for any non-empty set of loci $\{A_1,\ldots,A_n\}$, and any set of bits $\{b_1,\ldots,b_n\}$, one can construct a similar table by letting the marginal values of two genotypes $b_1\ldots b_n$, and $\overline{b_1}\ldots\overline{b_n}$ be some value $\delta$ and letting the marginal values of all other genotypes be $-\frac{2\delta}{2^n-2}$  (This observation will come in handy in our definition of type 1 pivotal functions in section \ref{pivotalugas}).

                                                                                                      If loci that interact also have statistically significant main effects then these loci will be detected by genome wide scans for main effects. Once detected, the interactions between the loci can be mapped. If, however, loci that interactively influence a quantitative trait have no main effects (or if their their main effects are statistically insignificant) then, as Frankel and Shork have explained, one will not detect such loci unless one explicitly uses an investigative technique that ``accommodates epistasis".

                                                                                                      Main effects can be detected by visiting loci one at a time and testing for differentiated marginals (marginals with non-zero marginal values). Let us call this strategy \emph{differentiated marginal testing} (DMT). To the best of our knowledge, the only sure way to accommodate for epistasis between loci when main effects are absent, is to visit \emph{multi-locus combinations}, and  to test the (multivariable) marginal of each such combination for differentiation. We shall call this approach \emph{combinatorial differentiated marginal testing}, or combinatorial DMT for short. The computational intractability of combinatorial DMT, even for small combination sizes, is discussed in a recent article by Moore \cite{moore}. Moore remarks:
                                                                                                      \begin{quote}``Identifying the optimal combination of [loci] from an astronomical number of possible combinations is computationally infeasible, especially when the [loci] do not have independent [i.e. main] effects. The following example illustrates the computational magnitude of the problem. Let's assume that $10^6$ [loci] have been measured. Let's also assume that 1,000 computational evaluations can be completed in one second on a single processor and that 1,000 processors are available for use. Exhaustively evaluating all of the approximately $4.9\times10^{11}$ two-way combinations of [loci] would require approximately 5.7 days. Exhaustively evaluating all of the approximately $1.6\times10^{17}$ three-way combinations of [loci]  would require 1,929,007 years. This of course assumes a best-case scenario in which the genetic model of interest consists of only two or three important attributes or genetic variations."\end{quote}

                                                                                                      \noindent The problem described above (see also \cite{moore2003une}, and \cite{moorejama}) is a specific instance of the general problem of identifying interacting attributes in data-mining \cite{freitas2001ucr}.

                                                                                                      In this paper we focus on generative versions of two specific problems having to do with the identification of interacting-attributes. Crucially,  main effects are entirely absent in both problems. By ``generative" we mean that the value of any synthesized data point can be queried---like in active learning. The first problem can only be solved by a combinatorial DMT strategy that tests attributes in combinations of two or more. The running time of such a strategy is therefore quadratic in the number of attributes. The second problem can only be solved by a combinatorial DMT strategy that tests attributes in combinations of four or more; the time required by this strategy is therefore  $\Omega(\ell^4)$, where $\ell$ is the number of attributes of an instance of the problem. We will show that both the first and the second problem can be solved robustly by an SGA in time that is \emph{linear} with respect to the number of attributes. Moreover, we will show that in both cases, the query complexity\footnote{Because fitness evaluation is by far the most time consuming part of a typical GA run, genetic algorithmicists often use the term \emph{time complexity} to refer to the relationship between a parameter of some problem and the number of fitness evaluations required by a GA to solve the problem. In this paper we use the term \emph{query complexity} instead. Our usage of this term is in line it's usage in theoretical computer science (the fitness function of a GA can be thought of as an oracle that gets ``queried" by the GA). We use the term \emph{time complexity} as it is typically used in computer science---to refer to the relationship between a problem parameter and the number of ``basic steps" required to solve the problem.} of the SGA is \emph{constant} with respect to the number of attributes.


                                                                                                      \section{Our Mode of Analysis} \label{modeofanalysis}

                                                                                                      This paper is somewhat unusual as foundational studies of genetic algorithms go in that experiments play a primary role in our mode analysis. Experiments are typically used in foundational GA research either to confirm behavior predicted by formal models, or to draw attention to phenomena not predicted by prevailing theories (e.g. \cite{Syswerda89,mitchell:1992:rrgaflgp,forrest93relative,mitchell1994wga}). The use of experiments as a primary tool of analysis is, however,  typically avoided because of the problem of \emph{specificity}.

                                                                                                      One can identify two kinds of specificity. First, as GAs are stochastic processes, any observations about the behavior of a GA during some run are, strictly speaking, only valid for the integer used to seed the random number generator. Of course, one can easily circumvent this problem by  running the GA several times with different seeds. Doing so allows one to build confidence that observed effects are not artifacts of some random seed. In most cases it is straightforward to quantify this confidence using statistics.

                                                                                                      The second kind of specificity is more problematic. Strictly speaking, an experimental result only pertains to the parameter values used in the experiment. In practice it may be possible, by changing a parameter while holding all others constant, to glean the relationship between that parameter and some aspect of GA behavior. However, if our aim is to be rigorous, then the extrapolation involved in this approach is less than ideal. In this paper we circumvent the problem with the second kind of specificity by exploiting the symmetries of the SGAs we construct. By doing we obtain hard quantitative results from a single experiment for an infinite set of problem instances.

                                                                                                      Symmetry arguments \cite{symmetry,jaynes}, while not new to GA research (for a previous instance see \cite{altenberg1997fdc}), are not common either. Such arguments are more frequently used in physics and chemistry. Indeed according to the theoretical  physicist E. T. Jaynes ``almost the only known exact results in atomic and nuclear structure are those which we can deduce by symmetry arguments, using the methods of group theory"\cite[p331-332]{jaynes}.

                                                                                                      One does not, however,  need to venture so far afield in order to find an example of a symmetry argument. Let $\mathfrak B_n$ denote the set of bitstrings of length $n$. For any bitstring $g$, let $\overline g$ denote the bitwise complement of $g$ (for example, $\overline{1011}=0100$). Let $\ell$ be some positive integer, and let $f$ be some fitness function over $\mathfrak B_\ell$ such that for any bitstring $g \in \mathfrak B_\ell$, $f(g)=f(\overline g)$ (for example, if $\ell=4$ and  $f(1011)=2.75$, then $f(0100)=2.75$). Let $G$ be some finite population SGA with fitness function $f$, such that the initial generation of $G$ is drawn from the uniform distribution over the set $\mathfrak B_\ell$.  For any generation $t$, let $p^{(t)}(g)$ denote the probability that some bitstring $g\in \mathfrak B_\ell$ will be in the population of $G$ in generation $t$. Then by appreciating the symmetry of the situation, we can deduce that for any generation $t$, and any bitstring $g\in \mathfrak B_\ell$, $p^{(t)}(g)=p^{(t)}(\overline g)$.

                                                                                                      This result holds regardless of the size of the population, the mutation and crossover rates, the mutation and crossover operators used, and the way in which the SGA $G$ scales the fitness values of individuals (if it does) and performs selection. Since mathematical models of genetic algorithms with finite populations  (e.g. \cite{nix1992mga}) tend to be unwieldy, a formal proof of the above, i.e. a proof within some formal axiomatic system, would be quite involved and would be relatively inaccessible. ``The great power of symmetry arguments lies just in the fact that they are not deterred by any amount of complication in the details", writes Jaynes  \cite[p331]{jaynes}.  Symmetry arguments, in other words,  allow one to cut through complications that might hobble other modes of argument.

                                                                                                      Jaynes stresses, as do we, that symmetry arguments rely not on `equal ignorance', but on `\emph{positive knowledge of symmetry}'. For instance, going back to the example above, if we cannot be sure that the initial population of $G$ is drawn from the uniform distribution over $\mathfrak B_\ell$,  then for any generation $t$, and any bitstring $g$, we would, in a sense, be `equally ignorant' of the values of $p^{(t)}(g)$ and $p^{(t)}(\overline g)$. Our `equal ignorance' does not, of course, entail that these two values are the same.

                                                                                                      But what constitutes `positive knowledge of symmetry'? This question, like the question ``what constitutes beauty?", has no direct answer. Historically, the appreciation of a symmetry by a community of theorists well versed in ``the art" was enough to constitute `positive knowledge' of that symmetry. Interestingly, over the last two centuries, mathematicians have largely agreed to eschew all non-sentential symmetries in their \emph{formal} communication; these days, only certain types of symmetry between sentential forms are acknowledged. At the beginning of this deep shift in the communication of mathematics, Euclidian geometry was one of the only fields of mathematics that had an axiomatic  foundation \cite{N27}. By the end of the shift, ``new as well old branches of mathematics \ldots were supplied with what appeared to be adequate sets of axioms" \cite{N27}. An obvious benefit of this shift is a reduction in the number of mistakes communicated. The rarely acknowledged cost is the impedance of timely and accessible communication of results derived through the insightful exploitation of  non-sentential forms of symmetry.

                                                                                                      \section{Type 1 Pivotal Functions}\label{pivotalugas}

                                                                                                      We begin by defining a class of fitness functions with bitstring inputs such that when any of these functions is queried with an infinite set of samples drawn from a uniform distribution over it's domain, no locus has a main effect, even though some loci may interact epistatically with others. For reasons that will soon become clear we call the members of this class \emph{pivotal functions}.

                                                                                                      For any positive integer $n$, let $[n]$ denote the set of positive integers $\{1,\ldots,n\}$. For any $n$-tuple $x$ and any $i\in[n]$ let $x_i$ denote the $i^{th}$ element of $x$. For any bitstring $s$ let $s_i$ denote the $i^{th}$ symbol of $s$. For any bit $b$ let $\overline{b}$ denote the complement of $b$. Let $\mathcal N(\mu,\sigma^2)$ denote the normal distribution with mean $\mu$ and variance $\sigma^2$.

                                                                                                      \begin{definition} \label{pivdisc} Let $\psi=(o,\sigma,\delta, \ell, L, V)$ be a 6-tuple such that $o$ is a positive integer,  $\sigma$ and $\delta$ are non-negative real numbers, $\ell$ is a positive integer greater than $o$, $V$ is an $o$-tuple of binary values, and $L$ is an $o$-tuple of distinct positive integers in $[\ell]$ sorted in ascending order. A \emph{type 1 pivotal function with descriptor} $\psi$ is a stochastic function $f$ over the set of bitstrings of length $\ell$ which behaves as follows: for any input bitstring $g$, if $(g_{L_1}=V_1 )\wedge\ldots\wedge (g_{L_o}=V_o)$ or $(g_{L_1}=\overline{V_1} )\wedge\ldots\wedge (g_{L_o}=\overline{V_o})$ then $f$ returns a value drawn from $\mathcal N(\delta, \sigma^2)$, otherwise $f$ returns a value drawn from $\mathcal N(-\frac{2\delta}{2^o-2},\sigma^2)$.
                                                                                                      \end{definition}

                                                                                                      We call $o,\delta,\sigma, \ell$ and $V$ the \emph{order, increment, noisiness, and span} of a pivotal function respectively.  When a pivotal function is queried with some bitstring, the distribution from which the result is drawn pivots upon the values of the bitstring at the \emph{pivotal loci} given by $L$, and the \emph{pivotal values} given by $V$---hence the name \emph{pivotal} function. If we assume that a type 1 pivotal function is queried with samples drawn from the uniform distribution over the function's domain, then the \emph{expected} marginal value of each allele of any individual locus is zero. This is what we mean when we say that no locus has a main effect. The increment parameter $\delta$ determines the strength of the expected multilocus marginal values of the pivotal loci.

\begin{figure}[t!]\begin{center}
                                                                                                      \includegraphics[width=.7\textwidth]{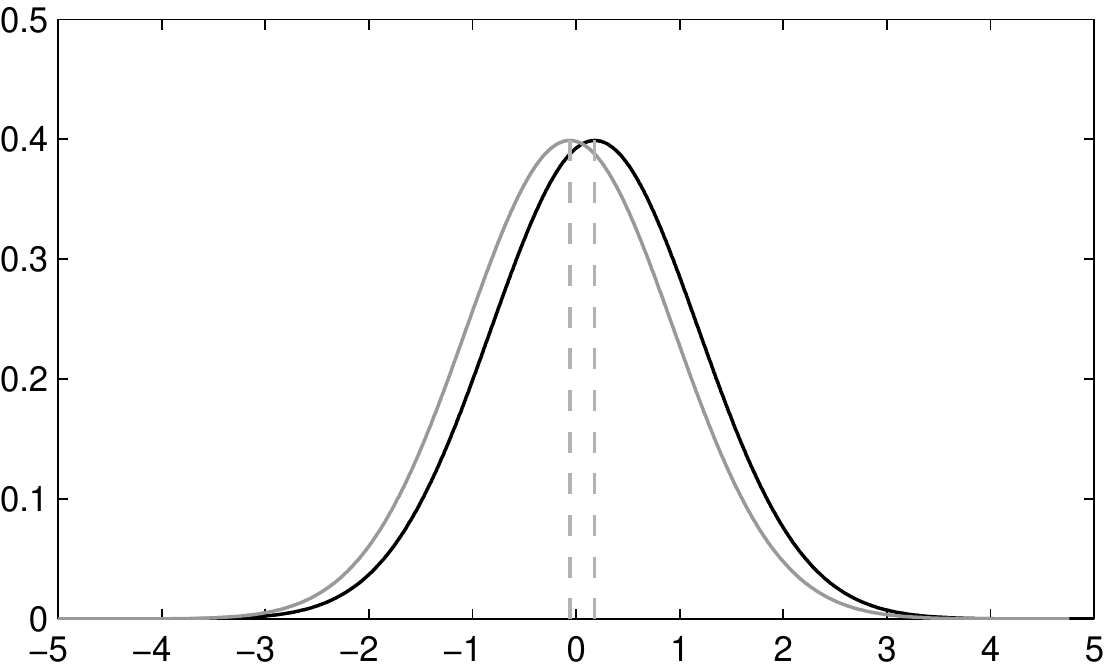}
                                                                                                      \caption{\label{gaussianpdfs}The pdfs of two normal distributions with standard deviation 1, and means -0.06 (grey) and 0.18 (black) }\end{center}
                                                                                                      \end{figure}

                                                                                                      Let $f$ be a type 1 pivotal function with descriptor $(o=3,\,\sigma=1, \,\delta=0.18,\, \ell,\, L,\, V)$. The pdfs  of $\mathcal N(-\frac{2\delta}{2^o-2}, \sigma^2)$ and $\mathcal N(\delta, \sigma^2)$ are shown in Figure \ref{gaussianpdfs}. Consider the task of robustly recovering the indices of the pivotal loci (i.e. the values of $L$) given only the values of $o,\sigma, \delta$ and $\ell$, and query access to the function $f$. Because of the stochastic nature of $f$, as long as there is any overlap between the two pdfs there will always be some probability of error. The \emph{large} overlap between the two pdfs shown in Figure \ref{gaussianpdfs} make the minimization of this error expensive as $\ell$ gets large (say $10^6$). But it is the absence of main effects that \emph{really} makes this problem thorny. A DMT strategy that visits loci one-by-one clearly will not work because none of the loci have main effects. Such a strategy only begins to hold promise if loci are visited in combinations of two or more. The number of such combinations however scales at least quadratically with $\ell$.

                                                                                                      We will show that an  SGA with uniform crossover can identify the pivotal loci  of $f$ relatively robustly (with less than a 0.005 chance of misclassification per locus) in time that is \emph{linear} in $\ell$, and with some number of queries that is \emph{constant} with respect to $\ell$.

                                                                                                      \subsection{Symmetry Analysis}

                                                                                                      For our purposes a \emph{semi-parameterized} SGA is an SGA with just two ``parameters": a positive integer $\ell$ which specifies the length of the genomes, and a fitness function over the set of bitstrings $\mathfrak B_\ell$. Two semi-parameterized SGAs are considered to be distinct if they differ in their crossover rates, say, or in the selection schemes that they use. It is important to clarify that the per-bit mutation rate of a semi-parameterized SGA is not dependent on the genome length. For any positive integer $\ell$, any semi-parameterized SGA  $G$ and any fitness function $f$ over $\mathfrak B_\ell$, we use the ``oracle notation" of theoretical computer science to denote the (unparameterized) SGA that results when the length of the bitstrings of $G$ is fixed at $\ell$, and the fitness function that $G$ queries is set to $f$; specifically, we denote this SGA by $G^f$.

                                                                                                    For any positive integer $n$ let $\Upsilon_n$ denote the set $\{0, \frac{1}{n}, \frac{2}{n},\ldots,\frac{n-1}{n},1\}$. Let us call the frequency of $1$'s and $0$'s in a population at some locus $k$ in some generation $t$ the \emph{one-frequency} and \emph{zero-frequency} respectively of locus $k$ in generation $t$. For any unparameterized SGA $A$ with population size $N$, let $\mathbf 1^{(t)}_{(A,i)}:\Upsilon_N\rightarrow [0,1]$ be a probability mass function such that, for any $x\in \Upsilon_N$, $\mathbf 1^{(t)}_{(A,i)}(x)$ is the probability that the one-frequency of locus $i$ after $t$ generations of running $A$ is $x$. Likewise let $\mathbf 0^{(t)}_{(A,i)}:\Upsilon_N\rightarrow [0,1]$ be a probability mass function such that, for any $x\in \Upsilon_N$, $\mathbf 0^{(t)}_{(A,i)}(x)$ is the probability that the zero-frequency of locus $i$ after $t$ generations of running $A$ is $x$. We call such distributions one- and zero-frequency distributions. Finally, let $\Uparrow^{(t)}_{(A,i)}$ and $\Downarrow^{(t)}_{(A,i)}$ be random variables that give the one- and zero-frequencies, respectively, of locus $i$ in generation $t$. Clearly then, the probability mass functions of $\Uparrow^{(t)}_{(A,i)}$, and $\Downarrow^{(t)}_{(A,i)}$ are $\mathbf 1^{(t)}_{(A,i)}$ and $\mathbf 0^{(t)}_{(A,i)}$ respectively. Note that for any $x\in\Upsilon_N$, $\mathbf 0^{(t)}_{(A,i)}(x)=\mathbf 1^{(t)}_{(A,i)}(1-x)$, and $\mathbf 1^{(t)}_{(A,i)}(x)=\mathbf 0^{(t)}_{(A,i)}(1-x)$.


                                                                                                      \begin{proposition} \label{oneequalszero} Let $G$ be a semi-parameterized  SGA, and let $f$ be a type 1 pivotal function. Then for any locus $k$ of $G^f$, and for any generation $t$
                                                                                                      \begin{itemize}
                                                                                                      \item[(a)] $\mathbf 1^{(t)}_{(G^f,k)}=\mathbf 0^{(t)}_{(G^f, k)}$
                                                                                                      \item[(b)] $\E{\Uparrow^{(t)}_{(G^f, k)}}=\E{\Downarrow^{(t)}_{(G^f, k)}}=\frac{1}{2}$
                                                                                                      \end{itemize}
                                                                                                      \end{proposition}

                                                                                                      \textsc{Argument:} For any generation $t$, and any locus $k$, Part (a) follows by consideration of the symmetry between $\mathbf 1^{(t)}_{(G^f,k)}$, and $\mathbf 0^{(t)}_{(G^f, k)}$ induced by the type 1 pivotal function $f$. Part (b) follows from part (a) and the claim that for any unparameterized SGA $A$, any locus $i$ of $A$, and for any generation $t$,  $\E{\Uparrow^{(t)}_{(A,i)}}+\E{\Downarrow^{(t)}_{(A,i)}}=1$. For a proof of the claim note that if $N$ is the size of the population of $G$, then for any generation $t$, and for any locus $i$,   \[\sum_{z\in\Upsilon_N}\z{t}{A,i}(z)z=\sum_{z\in\Upsilon_N}\z{t}{A,i}(1-z)(1-z)\] So,
                                                                                                      \begin{eqnarray*}&&\sum_{x\in\Upsilon_N}\o{t}{A,i}(x)x+\sum_{y\in\Upsilon_N}\z{t}{A,i}(y)y\\
                                                                                                      &&\phantom{aaa}=\sum_{x\in\Upsilon_N}\o{t}{A,i}(x)x+\z{t}{A,i}(1-x)(1-x)\\
                                                                                                      &&\phantom{aaa}= \sum_{x\in\Upsilon_N}\o{t}{A,i}(x)x+\o{t}{A,i}(x)(1-x)\\
                                                                                                      &&\phantom{aaa}=\sum_{x\in\Upsilon_N}\o{t}{A,i}(x)\\
                                                                                                      &&\phantom{aaa}=1 \end{eqnarray*}

                                                                                                      Note that proposition \ref{oneequalszero} holds for any type 1 pivotal function, and a semi-parameterized SGA with any population size, any commonly used selection operator (e.g. rank, tournament, fitness proportional etc.), any of the typical crossover and mutation operators, any mutation and crossover rates, and any fitness scaling scheme. Imagine having to prove all of this without appealing to the symmetries of SGAs with type 1 pivotal functions.

                                                                                                      Uniform crossover \cite{ackley1987cmg}, if present, adds yet another exploitable symmetry. This form of crossover was popularized by Syswerda \cite{Syswerda89}, who showed that uniform crossover can outperform one-point and two-point crossover on problems ranging from simple (e.g. one max) to complex  (the travelling salesperson problem). A large amount of evidence for the practical utility of uniform crossover has since accumulated. Syswerda also observed that \emph{any} homologous crossover operation can be represented by a probability distribution over the set of binary masks. Only in the case of uniform crossover, however, can the mask of a crossover operation be given by a string of  \emph{independent identically distributed} random binary variables. This  \emph{absence of positional bias} \cite{eshelman1989bcl}  is a crucial property of uniform crossover that we will exploit forthwith. For the sake of brevity we call an SGA with uniform crossover a UGA.

                                                                                                      \begin{definition}\label{standardform}
                                                                                                      Let $f$ be a type 1 pivotal function with descriptor $(o,\delta, \sigma, \ell, L, V)$. Then the \emph{basic form} of $f$ is a type 1 pivotal function with descriptor $(o,\delta,\sigma, o+1, (1,\ldots,o), (1,\ldots,1))$.
                                                                                                      \end{definition}

                                                                                                      According to this definition if some function $f^*$ is the basic form of some type 1 pivotal function $f$ with order $o$ then the span of $f^*$ is $o+1$. The first $o$ loci of any input to $f^*$ will be pivotal, the last locus of any input to $f^*$ will be non-pivotal, and the pivotal values used by  $f^*$ are all ones, We say that a type 1 pivotal function $f$ with descriptor $(o,\delta, \sigma, L,V)$ is \emph{basic} if the basic form of $f$ \emph{is} $f$. Since the last three elements of the descriptor of $f$ are then derivable from the first three, we write this descriptor as $(o,\delta,\sigma)$.

                                                                                                      \begin{proposition} \label{symmetry}
                                                                                                      Let $U$ be a semi-parameterized UGA, let $f$ be a type 1 pivotal function with descriptor $(o,\delta,\sigma,\ell, L, V)$, and let $f^*$ be the basic form of $f$. Then, for any generation $t$,

                                                                                                      \begin{itemize}
                                                                                                      \item[(a)] For any pivotal locus $k$,  $\mathbf 1^{(t)}_{(U^f,k)}=\mathbf 1^{(t)}_{(U^{f^*},1)}$
                                                                                                      \item[(b)] For any non-pivotal locus $k$, $\mathbf 1^{(t)}_{(U^f,k)}=\mathbf 1^{(t)}_{(U^{f^*},o+1)}$
                                                                                                      \end{itemize}
                                                                                                      \end{proposition}

                                                                                                      In other words the one-frequency distribution of any pivotal locus of $U^f$ in some generation $t$ is same as the one-frequency distribution of the first locus of $U^{f^*}$ in generation $t$, and the one-frequency distribution of any non-pivotal locus of $U^f$ in generation $t$ is the same as the one-frequency distribution of the last locus of $U^{f^*}$ in generation $t$.

                                                                                                      \noindent \textsc{Argument: }Let $f'$ be a type 1 pivotal function with descriptor $(o, \delta, \sigma,\ell, L, (1,\ldots,1))$ . We shortly present four claims. Part (a) of proposition \ref{symmetry} follows from claim \ref{GtoGsamelength}(a), claim \ref{pivcollate}, claim \ref{pivequal} and proposition \ref{oneequalszero}(a). Part (b) of the above proposition follows from claim \ref{GtoGsamelength}(b) and claim \ref{nonpivcollate}.

                                                                                                      \begin{claim} \label{GtoGsamelength}
                                                                                                       For any generation $t$, we have the following:
                                                                                                       \begin{itemize}
                                                                                                      \item[(a)] For any $i\in[o]$, if $V_{i}=1$, then $\mathbf 1^{(t)}_{(U^{f'},L_{i})}=\mathbf 1^{(t)}_{(U^f,L_{i})}$, otherwise $\mathbf 1^{(t)}_{(U^{f'},L_{i})}=\mathbf 0^{(t)}_{(U^f,L_{i})}$.

                                                                                                      \item[(b)]
                                                                                                      For any generation $t$, and any non-pivotal locus $k$ of $U^{f'}$, $\mathbf 1^{(t)}_{(U^{f'},k)}=\mathbf 1^{(t)}_{(U^f, k)}$
                                                                                                      \end{itemize}
                                                                                                      \end{claim}

                                                                                                      \begin{claim} \label{pivcollate}
                                                                                                      For any generation $t$, and any $i\in[o]$,$$\mathbf 1^{(t)}_{(U^{f^*},i)}=\mathbf 1^{(t)}_{(U^{f'},L_{i})}$$
                                                                                                      \end{claim}
                                                                                                      \begin{claim}\label{nonpivcollate}
                                                                                                      For any generation $t$, and any non-pivotal locus $k$ of $U^f$, $\mathbf 1^{(t)}_{(U^{f^*},o+1)}=\mathbf 1^{(t)}_{(U^{f'},k)}$
                                                                                                      \end{claim}

                                                                                                      \begin{claim} \label{pivequal}For any generation $t$, $$\mathbf 1^{(t)}_{(U^{f^*},1)}=\mathbf 1^{(t)}_{(U^{f^*},2)}=\ldots=\mathbf 1^{(t)}_{(U^{f^*},o)}$$
                                                                                                      \end{claim}

                                                                                                      Claim 1 follows from the observation that in any generation the population of $U^f$ can be ``changed into" the population of $U^{f'}$ and vice versa by a simple $0\leftrightarrow1$ relabeling of all genomic bits at those pivotal loci of $U$ whose corresponding pivotal values are  0.

                                                                                                      Claims \ref{pivcollate} follows by consideration of the symmetry between loci $L_1, \ldots, L_o$ of $U^{f'}$ and loci $1, \ldots, o$ of $U^{f^*}$ respectively. Claim \ref{nonpivcollate} follows by consideration of the symmetry between any non-pivotal locus of $U^{f'}$ and locus $o+1$ of $U^{f^*}$. These symmetries follows from the absence of positional bias in uniform crossover and from the definition of the ladder functions  $f^*$ and $f'$.

                                                                                                      \begin{figure*}[t!]\begin{center}
                                                                                                      \begin{ttfamily}\small
                                                                                                      \subfigure[``Vertical view" of a hypothetical population. Each \emph{row} is a genome]{\label{verticalselection}
                                                                                                      \renewcommand{\tabcolsep}{1pt}
                                                                                                      \begin{tabular}{r*{3}{>{\centering}p{16pt}|} >{\columncolor[gray]{.85}\centering}p{16pt}|*{3}{>{\centering}p{16pt}|}>{\columncolor[gray]{.85}\centering}p{16pt}|*{7}{>{\centering}p{16pt}|}>{\columncolor[gray]{.85}\centering}p{16pt}|*{1}{>{\centering}p{16pt}|}>{\centering}p{16pt}c}

                                                                                                      \phantom{$x =$}&1&0&1&1&0&1&0&0&1&0&1&0&1&0&0&0&1&0&\\
                                                                                                      \phantom{$m =$}&0&0&1&0&1&0&0&1&1&0&1&0&0&1&0&1&0&0&\\
                                                                                                      \phantom{$z =$}&1&0&0&1&0&0&1&1&1&0&1&0&1&1&0&0&1&0&\\
                                                                                                      &1&1&0&0&1&0&1&0&0&0&1&1&0&1&0&1&0&1&\\
                                                                                                      &1&0&0&0&0&1&0&1&0&0&1&0&0&1&0&1&1&0&\\
                                                                                                      &0&0&0&1&0&1&0&0&1&1&0&1&0&0&0&1&0&0&\\
                                                                                                      &1&1&0&1&0&0&1&1&1&0&1&1&0&1&0&0&1&0&\\

                                                                                                      &$\vdots$&$\vdots$&$\vdots$&$\vdots$&$\vdots$&$\vdots$&$\vdots$&$\vdots$&$\vdots$&$\vdots$&$\vdots$&$\vdots$&$\vdots$&$\vdots$&$\vdots$&$\vdots$&$\vdots$&$\vdots$&\\
                                                                                                      &$\vdots$&$\vdots$&$\vdots$&$\vdots$&$\vdots$&$\vdots$&$\vdots$&$\vdots$&$\vdots$&$\vdots$&$\vdots$&$\vdots$&$\vdots$&$\vdots$&$\vdots$&$\vdots$&$\vdots$&$\vdots$&\\
                                                                                                      &$\vdots$&$\vdots$&$\vdots$&$\vdots$&$\vdots$&$\vdots$&$\vdots$&$\vdots$&$\vdots$&$\vdots$&$\vdots$&$\vdots$&$\vdots$&$\vdots$&$\vdots$&$\vdots$&$\vdots$&$\vdots$&\\
                                                                                                      &$\vdots$&$\vdots$&$\vdots$&$\vdots$&$\vdots$&$\vdots$&$\vdots$&$\vdots$&$\vdots$&$\vdots$&$\vdots$&$\vdots$&$\vdots$&$\vdots$&$\vdots$&$\vdots$&$\vdots$&$\vdots$&\\
                                                                                                      \end{tabular}}
                                                                                                      \subfigure[``Vertical view" of a hypothetical crossover operation]{\label{verticalcrossover}
                                                                                                      \renewcommand{\tabcolsep}{1pt}

                                                                                                      \begin{tabular}{r*{3}{>{\centering}p{16pt}|} >{\columncolor[gray]{.85}\arrayrulecolor{black}\centering}p{16pt}|*{3}{>{\centering}p{16pt}|}>{\columncolor[gray]{.85}\centering}p{16pt}|*{7}{>{\centering}p{16pt}|}>{\columncolor[gray]{.85}\centering}p{16pt}|*{1}{>{\centering}p{16pt}|}>{\centering}p{16pt}c}

                                                                                                      $x =$&1&0&0&1&0&0&1&1&1&0&1&0&1&1&0&0&1&0&\\
                                                                                                      $m =$&$X_1$&$X_2$&$X_3$&$X_4$&$X_5$&$X_6$&$X_7$&$X_8$&$X_9$&$X_{10}$&$X_{11}$&$X_{12}$&$X_{13}$&$X_{14}$&$X_{16}$&$X_{16}$&$X_{17}$&$X_{18}$&\\
                                                                                                      $y =$&1&1&0&0&1&0&1&0&0&0&1&1&0&1&0&1&0&1&\\
                                                                                                      \cline{2-19}
                                                                                                      \addlinespace[.05em]
                                                                                                      $z=$&?&?&?&?&?&?&?&?&?&?&?&?&?&?&?&?&?&?&\\

                                                                                                      \end{tabular}}

                                                                                                      \end{ttfamily}
                                                                                                      \end{center}
                                                                                                      \caption{Subfigure (a) shows a ``vertical view" of a hypothetical population. Each row is a genome. The shaded columns show the positions of three hypothetical pivotal loci. By the definition of a type 1 pivotal function (see text), only the bits in the shaded columns matter during selection. Subfigure (b) shows a ``vertical view" of a hypothetical crossover operation. Two parents, $x$ and $y$ are about to undergo uniform crossover which will yield a child $z$. The bits of  $z$ will be determined by the values of the independent identically distributed random binary variables that comprise the mask $m$.\label{vertical}}

                                                                                                      \end{figure*}
                                                                                                      To make these symmetries manifest, we offer the two ``vertical views" shown in Figure \ref{vertical}. Figure \ref{verticalselection} shows a hypothetical population of $U^{f'}$ with three pivotal loci (whose locations are marked by shaded columns). Given the definition of the fitness function of $U^{f'}$, it is easy to see that the fitness of any genome depends only upon the value of that genome's bits at the pivotal loci. Thus only the bits in the shaded columns of Figure \ref{verticalselection} matter in determining a genome's fitness, and by extension its chance of being selected  (note that this is true regardless of the selection scheme used). Figure \ref{verticalcrossover} shows a ``vertical view" of a hypothetical uniform crossover operation in $U^{f'}$. Two genomes, $x$ and $y$,  have been selected for uniform crossover. The crossover mask $m$ is represented as a string of random binary variables. The values of these variables determines the bits of the child $z$. Because crossover is uniform, the random variables in $m$ are independent and identically distributed.

                                                                                                      Claim \ref{pivequal} follows from the symmetry that exists between each of the first $o$ loci of $U^{f^*}$. This symmetry follows from the absence of positional bias in uniform crossover, and from the definition of the fitness function used by $U^{f^*}$. $\Box$

                                                                                                      By proposition \ref{symmetry}(a), for any pivotal locus $k$ of $U^f$, drawing monte-carlo samples from $\mathbf 1^{(t)}_{(U^{f*},1)}$ is equivalent to drawing monte-carlo samples from $\mathbf 1^{(t)}_{(U^f,k)}$. And by proposition \ref{symmetry}(b), for any non-pivotal locus $k$ of $U^f$, drawing monte-carlo samples from $\mathbf 1^{(t)}_{(U^{f^*},o+1)}$ is equivalent to drawing monte-carlo samples from $\mathbf 1^{(t)}_{(U^f,k)}$.

                                                                                                      \subsection{Experiment 1}\label{anexperiment}

\begin{figure*}
                                                                                                      \subfigure{\includegraphics[width=.5\textwidth]{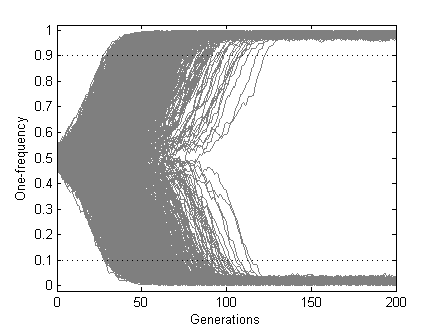}}
                                                                                                      \subfigure{\includegraphics[width=.5\textwidth]{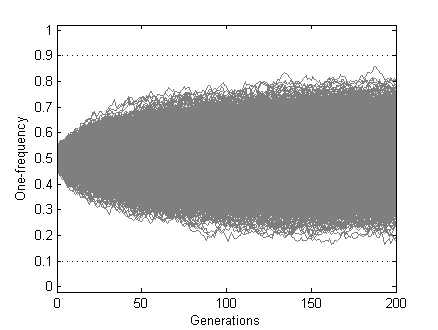}}
                                                                                                      \caption{\label{initialexperiment} The one-frequency dynamics of the first (\emph{left}) and fourth (\emph{right}) loci of the UGA $W^{f^*_1}$}
                                                                                                      \end{figure*}

                                                                                                      Let $W$ denote the semi-parameterized UGA described in the materials and methods section in the appendix, and let $f_1^*$ denote a basic type 1 pivotal function with descriptor $(o=3,\,\delta=0.18,\,\sigma=1)$. Figure \ref{initialexperiment} shows the one-frequency dynamics of the first and last locus of $U^{f^*}$  in each of 3000 runs. In all 3000 runs the first locus went to fixation\footnote{We use the term `fixation' loosely. Clearly, as long as the mutation rate is non zero, no locus can ever be said to go to fixation in the strict sense of the word.} by the 200th generation, whereas the one-frequency of the last locus in the 200th generation was always between 0.93 and 0.07.

                                                                                                      In order to clearly describe the rest of our findings we  develop the following notation.  We denote a \emph{schema partition} \cite{Mitchell:1996:IGA} by a tuple consisting of the indices of the defining positions of that schema partition---e.g. $(2, 15, 3)$. The \emph{order} of a schema partition $\Gamma$, denoted by $o(\Gamma)$, is the number of elements in some tuple that denotes $\Gamma$. The denotation of a \emph{schema} is dependent on the denotation of the schema-partition that the schema belongs to. For any genome $g$, let $g_i$ denote the $i^{th}$ bit of $g$. Given a schema partition denoted by some tuple $\Gamma$, the schemata in this partition are denoted by binary strings of length  $o(\Gamma)$.  Let $b_1,\ldots,b_{o(\Gamma)}$ be some bits. Then,  $b_1\ldots b_{o(\Gamma)}$ denotes the schema consisting of the genomes $ \{g|g_1=b_1\wedge\ldots\wedge g_{o(\Gamma)}=b_{o(\Gamma)}\}$. The denotation of the relevant schema partition must always be borne in mind when interpreting a denoted schema..

                                                                                                      In addition to the findings reported above, we found that 200 generations into each run of $W{f_1^*}$, either the schema 000, or the schema 111, of the schema partition $( 1,2,3)$, dominated the population. The average fraction of the population that belonged to the dominant schema at the end of 200 generations was 0.9563 (with standard error $1.56\times10^{-4}$).



                                                                                                      Given the conclusions of our symmetry analysis, the result shown in Figure \ref{initialexperiment} provides us with a window into the frequency dynamics of any UGA $W^{f_1}$, where $f_1$ is a type 1 pivotal function whose basic form is $f^*_1$.  We infer that the pivotal loci of $W^{f_1}$ will tend to go to fixation by the 200th generation. We also infer that the divergence from $0.5$ of the one-frequencies of the non-pivotal loci of $W^{f_1}$ will tend not to be not extreme.

                                                                                                      We now explain the behavior of $W^{f_1}$ that we have just deduced. Note that while this discussion is speculative and imprecise, it is entirely tangential to our aim of identifying computational competencies of the SGA.  The one-frequency dynamics of the non-pivotal loci of $W^{f_1}$ is easily explained by the notion of \emph{drift}. To understand the frequency dynamics of the pivotal loci, it helps to go back to the ``vertical views" presented in Figure \ref{vertical}. Observe that discounting the effect of sampling error, only selection and mutation have an effect on the composition of the bit-pool of each locus\footnote{Changes in the one and zero frequencies of a locus can be visualized as changes in the composition of a \emph{pool of bits}. The bit-pool metaphor is especially useful in conjunction with the ``vertical view" of a population presented in figure \ref{verticalselection}; each column can be thought of as a pool of bits}. Crucially, crossover does not change the composition of these bit-pools. Now, let $x_1, x_2,$ and $x_3$ denote the indices of the pivotal loci of $W^{f_1}$, and, without loss of generality, suppose that the pivotal values of the first, second, and third pivotal loci are 0, 1, and 0 respectively. Consider the frequency dynamics of the schema 010 of the site-array $\langle x_1, x_2, x_3\rangle$. The probability of generating genomes of type 010 in some generation is highly (though not completely) dependent upon the composition of the bit-pools of the pivotal loci in the previous generation. A genome of type 010, once generated, will tend to be preferentially selected over all other genomes except those that belong to the ``sibling" schema 101. Thus, regardless of what happens during crossover, once generated, a genome of type 010 will tend to increase the frequency of 0, 1, and 0 in the bit-pools of the first, second, and third pivotal loci respectively. This makes conditions more favorable for the generation of genomes of type 010 in future generations. Of course, the same argument applies to genomes of type 101. Now, given that the alleles 1 and 0 are, in a sense,``rivals" of each other at each locus, the schemata 010 and 101 ``compete" for dominance of the bit-pools of each of the pivotal loci. One of these schemata eventually manages to gain an edge in ``pulling" the composition of the bit-pools of all three pivotal loci far enough in it's favor that a self-reinforcing loop that heavily favors the future generation of the victorious schema then ensues.

                                                                                                      In light of this analysis, one can conclude that the building block hypothesis \cite{Goldberg:1989:GAS,Mitchell:1996:IGA,journals/ec/Holland00} takes an overly-grim view of the disruption of fit low-order schemata with high defining-lengths. This view misses the fact that the ``debris" from the disruption of such a schema changes the composition of the bit-pools at the defining positions of that schema in a way that favors the future generation of genomes belonging to the schema.

                                                                                                      \begin{algorithm}[tbp]\small
                                                                                                      \dontprintsemicolon
                                                                                                      \KwIn{\mbox{a type 1 or type 2 pivotal function $f$ }}
                                                                                                      $\ell$=span of $f$\;
                                                                                                      $pivotalLoci=$\{\}\;
                                                                                                      $nonPivotalLoci=$\{\}\;
                                                                                                      $P=$ population of  $W^f$ after $n$ generations\;
                                                                                                      \For{$i=1$ to $\ell$}{
                                                                                                      $x$ = one-frequency of locus $i$ in population $P$\;
                                                                                                      \eIf{$0.1 \leq x$ and $x \leq 0.9$}
                                                                                                      {$nonpivotalLoci=nonPivotalLoci$ $\cup$ $\{i\}$\;}
                                                                                                      {$pivotalLoci=pivotalLoci$ $\cup$ $\{i\}$\;}}
                                                                                                      \KwRet $pivotalLoci, nonPivotalLoc$i\;
                                                                                                      \caption{\textsc{ClassifyLoci}$\_n$}
                                                                                                      \end{algorithm}
                                                        %

                                                                                                     \subsection{A Computational  Competency of the SGA} \label{computationalstrength}

                                                                                                      Consider Algorithm 1. The results of our experiment with $W^{f^*}$ suggest that when $\textsc{ClassifyLoci}\_200$ is applied to $f_1$, it will classify each locus of $f_1$ fairly accurately. Let us quantify this accuracy. Note that in \emph{all} 3000 runs of $W^{f_1^*}$, by the end of the 200th generation, the one-frequency of the first locus was outside the interval [0.1, 0.9], and the one-frequency of the last locus was inside this interval. Let $a$ be the probability that the one-frequency of the first locus of $W^{f_1^*}$ will be \emph{inside} [0.1 0.9] at the end of the 200th generation. Let $H_0$ be the hypothesis that $a\geq 0.005$. If $H_0$ is true then the probability that the one-frequency of the first locus will be outside [0.1, 0.9] at the end of the 200th generation in \emph{each} of 3000 runs (as observed in the above experiment) is less than $(1-0.005)^{3000}<3\times 10^{-7}$. Therefore we reject $H_0$ at the $3\times10^{-7}$ level of significance. Now consider the hypothesis that with probability greater than or equal to 0.005 the one-frequency of the last locus of $W^{f_1^*}$ will be \emph{outside} the interval [0.1,0.9] at the end of 200 generations. Using very similar reasoning we reject this hypothesis at the $3\times10^{-7}$ level of significance. Thus, with probability of error less than three in ten million, the following statement is true: for any locus $k$ of $f_1$, there is less than a 0.005 probability that \textsc{ClassifyLoci}$\_200$ will misclassify locus $k$.

                                                                                                      Note that  $\ell$, may be \emph{any} positive integer greater than $3$. There are ${ \ell\choose 3}\in \Omega(\ell^3)$ possible combinations of the three pivotal indices\footnote{To obtain this bound we have used the inequality $\left(n/k\right)^k\leq{n\choose k}$. See \cite{clr}.}. Remarkably, \textsc{ClassifyLoci}$\_200$ achieves the level of robustness mentioned above ($p<0.005$ per locus) in time that is linear in $\ell$, after making some number of fitness evaluations that is \emph{constant} with respect to $\ell$

                                                                                                      \section{Type 2 Pivotal Functions} \label{typeIIpivotalsgas}

                                                                                                      If the order $o$ of a type 1 pivotal function is greater than one, then even though no individual locus will have differentiated marginal effects, certain combinations of \emph{two} loci, \emph{will} have differentiated (multilocus) marginal effects (specifically those combinations in which both loci are pivotal). Thus for any $o>1$, loci can be classified as pivotal or non-pivotal with a fixed level of robustness using a combinatorial DMT strategy that visits some number of two-locus combinations that is quadratic in the span of the pivotal function\footnote{As $o$ increases, the constant associated with this scaling relationship will increase very quickly. Nevertheless for any fixed value of $o$, the number of combinations that must be visited scales quadratically with the span of a type 1 pivotal function.}. Type 2 pivotal functions are expressly defined so that for any even order $o$, and any positive integer $m<o$, no combination of $m$ loci will have differentiated marginal effects!

                                                                                                      Let $\oplus$ denote the exclusive-or  operator (also the binary addition modulo 2 operator). Type 2 pivotal functions are defined as follows:

                                                                                                      \begin{definition} \label{piv2disc} Let $\psi=(o,\sigma,\delta, \ell, L)$ be a 5-tuple such that $o$ is a positive even integer,  $\sigma$ and $\delta$ are positive real numbers, $\ell$ is a positive integer greater than $o$, and $L$ is an $o$-tuple of positive integers in $[\ell]$ sorted in ascending order. A \emph{type 2 pivotal function with descriptor} $\psi$ is a stochastic function $f$ which behaves as follows: for any input bitstring $g$, if $g_{L_1}\oplus\ldots\oplus g_{L_o}=1$ then $f$ returns a value drawn from $\mathcal N(\delta, \sigma^2)$, otherwise $f$ returns a value drawn from $\mathcal N(-\delta,\sigma^2)$.
                                                                                                      \end{definition}

                                                                                                      The order of a type 2 pivotal function is always even; furthermore, no pivotal values are associated with the pivotal loci.

                                                                                                      Let $f$ be some type 2 pivotal function with span $\ell$. Observe that for any bitstring $g$ of length $\ell$,  $f(g)=f(\overline{g})$. Observe also that as $\oplus$ is associative and commutative, the order in which it is applied to the pivotal bits of some bitstring is immaterial. Both of these observations reveal symmetries of $f$ that we will exploit forthwith. These symmetries can also be seen in Table \ref{xortable}, which shows the marginal of the pivotal loci of some type 2 pivotal function with increment $\delta$, and order four.

                                                                                                      Finally observe that for any type 2 pivotal function with order $o$, the multilocus marginal  of any combination of $m<o$ alleles of distinct loci will not be differentiated. Thus the time complexity of a combinatorial DMT strategy that robustly identifies the pivotal loci is  $\Omega(\ell^o)$.

                                                                                                      Let $f$ be a type 2 pivotal function with descriptor $(o=4,\,\delta=0.25,\,\sigma=1,\,\ell, \,L)$. We now show that a UGA can identify the pivotal loci  of $f$ relatively robustly (with less than a 0.005 chance of misclassification per locus) in time that is \emph{linear} in $\ell$, and with some number of queries that is \emph{constant} with respect to $\ell$. Our approach is almost identical to the approach we took in Section \ref{pivotalugas}, where we showed a similar result for a class of pivotal functions of type 1.

                                                                                                      \begin{table}[t!]
                                                                                                      \begin{center}
                                                                                                      \begin{tabular}{cccc|c}\multirow{2}{*}{$A$}&\multirow{2}{*}{$B$}&\multirow{2}{*}{$C$}&\multirow{2}{*}{$D$}&Expected\\
                                                                                                      &&&&Marginal Values
                                                                                                      \\\hline 0&0&0&0&$-\delta$
                                                                                                      \\0&0&0&1&$+\delta$\\
                                                                                                      0&0&1&0&$+\delta$\\
                                                                                                      0&0&1&1&$-\delta$\\
                                                                                                      0&1&0&0&$+\delta$\\
                                                                                                      0&1&0&1&$-\delta$\\
                                                                                                      0&1&1&0&$-\delta$\\
                                                                                                      0&1&1&1&$+\delta$\\
                                                                                                      1&0&0&0&$+\delta$\\
                                                                                                      1&0&0&1&$-\delta$\\
                                                                                                      1&0&1&0&$-\delta$\\
                                                                                                      1&0&1&1&$+\delta$\\
                                                                                                      1&1&0&0&$-\delta$\\
                                                                                                      1&1&0&1&$+\delta$\\
                                                                                                      1&1&1&0&$+\delta$\\
                                                                                                      1&1&1&1&$-\delta$\end{tabular}
                                                                                                      \end{center}
                                                                                                      \caption{\label{xortable}The expected marginal of the pivotal loci of a type 2 pivotal function}
                                                                                                      \end{table}

                                                                                                      \subsection{Symmetry Analysis}

                                                                                                      We define the \emph{basic form} of a type 2 pivotal function as follows:

                                                                                                      \begin{definition}\label{standardform}
                                                                                                      Let $f$ be some type 2 pivotal function with descriptor $(o,\delta, \sigma, \ell, L)$. We define the basic form of $f$ to be a type 2 pivotal function with descriptor $(o,\delta,\sigma, o+1, (1,\ldots,o))$ \end{definition}

                                                                                                      Let $f$ be a type II pivotal function with descriptor $(o,\delta,\sigma,\ell,L,V)$. We say that $f$ is \emph{basic} if the basic form of $f$is $f$. Since the last two elements of  the descriptor of $f$ are derivable from the first three, we write this descriptor as $(o,\delta,\sigma)$.

                                                                                                      \begin{proposition} \label{symmetry2}
                                                                                                      Let $U$ be a semi-parameterized UGA, let $f$ be a type 2 pivotal function with descriptor $(o,\delta,\sigma,\ell, L, V)$, and let $f^*$ be the basic form of $f$. Then, for any generation $t$,

                                                                                                      \begin{itemize}
                                                                                                      \item[(a)] For any pivotal locus $k$,  $\mathbf 1^{(t)}_{(U^f,k)}=\mathbf 1^{(t)}_{(U^{f^*},1)}$
                                                                                                      \item[(b)] For any non-pivotal locus $k$, $\mathbf 1^{(t)}_{(U^f,k)}=\mathbf 1^{(t)}_{(U^{f^*},o+1)}$
                                                                                                      \end{itemize}
                                                                                                      \end{proposition}

                                                                                                      This proposition is almost word-for-word identical to Proposition \ref{symmetry}. Likewise the argument for this proposition, is very similar to the argument for Proposition \ref{symmetry}. We omit this argument on the assumption that it will be clear to readers who have digested the argument for Proposition \ref{symmetry}.

                                                                                                      By proposition \ref{symmetry2}(a), for any pivotal locus $k$ of $U^f$, drawing monte-carlo samples from $\mathbf 1^{(t)}_{(U^{f*},1)}$ is equivalent to drawing monte-carlo samples from $\mathbf 1^{(t)}_{(U^f,k)}$. And by proposition \ref{symmetry2}(b), for any non-pivotal locus $k$ of $U^f$, drawing monte-carlo samples from $\mathbf 1^{(t)}_{(U^{f^*},o+1)}$ is equivalent to drawing monte-carlo samples from $\mathbf 1^{(t)}_{(U^f,k)}$.

                                                                                                      \subsection{Experiment 2}\label{experiment2}
\begin{figure*}
                                                                                                      \subfigure{\includegraphics[width=.5\textwidth]{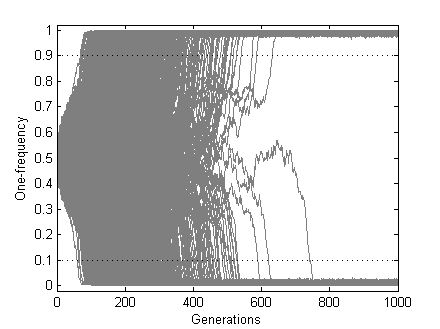}}
                                                                                                      \subfigure{\includegraphics[width=.5\textwidth]{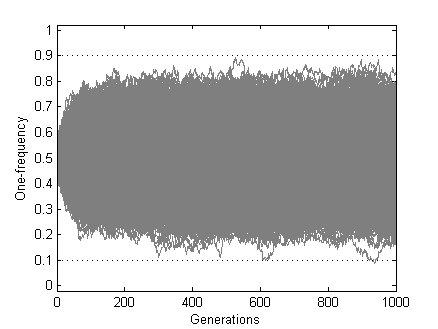}}
                                                                                                      \caption{\label{secondexperiment} The one-frequency dynamics of the first (\emph{left}) and fifth (\emph{right}) loci of the UGA $W^{f^*_2}$}
                                                                                                      \end{figure*}

                                                                                                      Recall that $W$ denotes the semi-parameterized UGA described in the materials and methods section in the appendix. Let  $f^*_2$ denote a basic type 2 pivotal function with descriptor $(o=4, \,\delta=0.25,\,\sigma=1)$. We executed 3000 runs of the UGA $W^{f^*_2}$. The one-frequency dynamics of the first and last loci in each run is plotted in Figure \ref{secondexperiment}. In all 3000 runs the first locus went to fixation by generation 1000, whereas the one-frequency of the last locus in generation 1000 was always between 0.9 and 0.1. We found that 1000 generations into each run the population was dominated by some schema $b_1b_2b_3b_4$ of the site-array $(1,2,3,4)$ with  $b_1\oplus b_2\oplus b_3\oplus b_4=1$. On average the fraction of the population that belonged to the dominant schema in generation 1000 was 0.9634 (with standard error $1.22\times 10^{-4}$).

                                                                                                      Let $f_2$ be a type 2 pivotal function with span  $\ell$ such that the basic form of $f_2$ is $f^*_2$. The conclusions of our symmetry analysis of type 2 pivotal fitness functions, and the result shown  in Figure \ref{secondexperiment} provide us with a window into the frequency dynamics of all pivotal and non-pivotal loci of  $W^{f_2}$.

                                                                                                      \subsection{A Second Computational Competency of the SGA}

                                                                                                      Based on arguments that are almost identical to the ones in section \ref{computationalstrength} we conclude, with probability of error less than three in ten million, that the following statement is true: For any locus of $f_2$, the probability that the locus will be misclassified by \textsc{ClassifyLoci}$\_1000$  is less than a 0.005.
                                                                                                      There are ${ \ell\choose 4}\in \Omega(\ell^4)$ possible configurations of the pivotal indices. Remarkably, \textsc{ClassifyLoci}$\_1000$ achieves the level of robustness mentioned above ($p<0.005$ per locus) in time that is linear in $\ell$, after making some number of fitness evaluations that is \emph{constant} with respect to $\ell$.

                                                                                                      It merits mentioning that both the computational competencies showcased in this paper are essentially invisible to analytic approaches in which an infinite population is assumed (e.g. \cite{VoseLiepins91,interPopConstaints,CoarseGrainingFoga2007}). This is because without some kind of symmetry breaking, the one and zero frequencies of the pivotal loci  will not depart from 1/2. The role of symmetry breaking is performed here by sampling error which is absent in infinite population models of genetic algorithms.

                                                                                                      \section{Conclusion}\label{conclusion}

                                                                                                      This paper can be viewed as a response of sorts to the nihilism that the no free lunch theorems  \cite{ieee-ec:Wolpert+Macready:1997} have inspired. As mentioned in the introduction, no free lunch theorems are frequently regarded as ``proof" that in the \emph{practice} of black-box optimization, one search algorithm is as good as another. If this is indeed the case,  then genetic algorithms have no special advantage over other search algorithms; the fortuitous paring of optimization problem with search algorithm is all. Effort toward the identification of computational efficiencies underlying the genetic algorithm's capacity for practical blackbox optimization then seems misplaced.

                                                                                                      Nihilism of this sort is puzzling given the frequency with which genetic algorithms are successfully applied as black-box optimization algorithms in practice. The importance accorded to Wolpert and Mcready's results seems to be a reaction, or rather, an overreaction, to the empirical discovery of serious flaws  \cite{mitchell:1992:rrgaflgp,forrest93relative,mitchell1994wga}  in certain widely accepted statements about the computational efficiency of the simple genetic algorithm; these statements came to be regarded as fact because of an unclear demarcation between mathematical deduction and speculation in the early genetic algorithmics literature \cite{DBLP:journals/corr/abs-0810-3356}.

                                                                                                      It is important to emphasize that the problem was not speculation itself but the \emph{blurring} of the boundary between deduction and speculation. If the mistakes of the past are not to be revisited, this boundary must be clearly maintained at all times. Accordingly, we have distinguished between a computational \emph{competency} of the SGA---a  \emph{specific} computational efficiency that can be derived deductively\footnote{We clarify here that we offer the symmetry arguments presented in this paper as \emph{deductive} arguments, not speculative ones. These results have, of course, not been derived within some formal axiomatic system. However, to argue that this automatically makes our arguments non-deductive is to argue that deduction played a very limited role in mathematics before the nineteenth century.}---and a computational \emph{proficiency} of the SGA---a \emph{general} computational efficiency that is inferred inductively from a set of related computational competencies.

                                                                                                      In this paper we have derived two closely related computational competencies of the SGA. The general domain in which these competencies lie is noteworthy. A central explanandum of the field of genetic algorithmics, and indeed, of the field of evolutionary biology, is the persistence of adaptation in sexually evolving populations \emph{despite} the ubiquity of epistatic interactions between unlinked genomic loci. That the SGA can, in particular cases, robustly and scalably ``identify" small numbers of unlinked epistatically interacting loci \emph{with no main effects}, and moreover, that the SGA does so by sending specific genotypes with above average fitness to fixation, is likely to be material to theories about the adaptive prowess of both the genetic algorithm and natural evolution.

                                                                                                      \bibliographystyle{plain}
                                                                                                      \bibliography{c:/mystuff/mycreations/0Work/refs}
                                                                                                      \appendix
                                                                                                      \section*{Materials and Methods}

                                                                                                      The arguments in this paper rest not just on the qualitative behavior of the semi-parameterized UGA we used, but on aspects of it's \emph{quantitative} behavior. It is therefore necessary to describe this UGA in sufficient detail that these aspects can be reproduced. While we believe that the description given below is sufficient to reproduce all quantitative aspects of the behavior of our UGA that are relevant to our proofs, small differences in implementation may interfere with reproducibility. We therefore release the Matlab code for the semi-parameterized UGA that we used. This code, along with the code for the pivotal fitness functions used is available for download\footnote{\url{http://cs.brandeis.edu/~kekib/competenciesMatlab.zip}}.

                                                                                                      The semi-parameterized UGA we used (denoted by $W$ in this paper) implements the specification for a simple genetic algorithm given by Mitchell \cite[p 10]{Mitchell:1996:IGA}, with two exceptions:
                                                                                                         \begin{enumerate}
                                                                                                         \item In each generation, right after evaluating the fitness of all individuals, our UGA used sigma scaling \cite[p 167]{Mitchell:1996:IGA} to adjust the fitness of each individual, and used this adjusted fitness when selecting the parents of that generation. Suppose $f^{(t)}_x$ is the fitness of some individual $x$ in some generation $t$, and suppose the average fitness and standard deviation of the fitness of the individuals in generation $t$ are given by $\overline{f^{(t)}}$ and  $\sigma^{(t)}$ respectively, then the adjusted fitness of $x$ in generation $t$ is given by $h^{(t)}_x$ where, if $\sigma^{(t)}=0$ then $h^{(t)}_x=1$, otherwise,  $$h^{(t)}_x=\min(0,1+\frac{f^{(t)}_x -\overline{f^{(t)}}}{\sigma^{(t)}})$$
                                                                                                         \item The SGA used universal stochastic stochastic sampling \cite{Baker:1985:ASM} \cite[p 166]{Mitchell:1996:IGA} to select parents.
                                                                                                         \end{enumerate}

                                                                                                      Selection is fitness-proportionate. the population size is 1500,  the probability of crossover is one,  and bit-flip mutation with a mutation rate of $0.003$ per bit is used.

                                                                                                      \end{document}